\documentclass{article}

\usepackage{arxiv}

\usepackage[utf8]{inputenc} 
\usepackage[T1]{fontenc}    
\usepackage{hyperref}       
\usepackage{url}            
\usepackage{booktabs}       
\usepackage{amsfonts}       
\usepackage{nicefrac}       
\usepackage{microtype}      
\usepackage{amsmath}        
\usepackage{lipsum}
\usepackage{graphicx}
\usepackage{float}
\graphicspath{ {./images/} }
\usepackage{xcolor}

\usepackage{tikz}
\usetikzlibrary{shapes.geometric, arrows.meta, positioning, calc, backgrounds, fit, decorations.pathreplacing}
\usepackage{amsmath, amssymb}

\definecolor{stage1blue}{RGB}{230, 242, 250}
\definecolor{stage2green}{RGB}{232, 245, 233}
\definecolor{stage3dark}{RGB}{38, 70, 83}
\definecolor{stage4orange}{RGB}{255, 243, 224}
\definecolor{headerblue}{RGB}{52, 152, 219}
\definecolor{cliffred}{RGB}{192, 57, 43}

\begin{document}

\title{Lost in Sampling: Assessing Lexical Reachability in LLMs via the Word Coverage Score (WCS)}

\author{
  Samer Awad, Javier Conde, Carlos Arriaga \\
  Information and Processing Telecommunications Center \\
  Universidad Politécnica de Madrid \\
  Madrid, Spain \\
  \And
  Tairan Fu \\
  Politecnico di Milano \\
  Milano, Italy \\
  \And
  Javier Coronado-Blázquez \\
  Banco de España \\
  Madrid, Spain \\
  \And
  Pedro Reviriego \\
  Information and Processing Telecommunications Center \\
  Universidad Politécnica de Madrid \\
  Madrid, Spain \\
}

\maketitle
\begin{abstract}
Modern Large Language Models (LLMs) are often criticized for producing repetitive and homogeneous text, despite possessing vast latent vocabularies. While previous research has focused on model knowledge and training data, we investigate the role of decoding mechanics in suppressing linguistic diversity. We introduce the Word Coverage Score (WCS), a metric that quantifies the extent to which contextually appropriate human vocabulary is mathematically pruned by standard sampling filters (e.g., Top-$p$, Top-$k$, and Min-$p$). Rather than assessing static knowledge, the WCS measures the lexical survival rate of low-frequency, high-information human words as a function of sampling parameters. By auditing open-weight models on human-authored corpus fragments, we identify which logical lexical choices are rendered unreachable by the decoder, even when they reside within the probability space. Our results provide quantitative evidence that industry-standard sampling defaults act as unintended censorship mechanisms, smoothing the unique textures of human expression into a homogenized discourse. The WCS offers a rigorous framework for optimizing the trade-off between text coherence and lexical richness, providing a diagnostic tool for preserving the diversity of human language in generative models.


\end{abstract}


\section{Introduction}
\label{sec:introduction}

The texts generated by Large Language Models (LLMs) tend to be homogeneous, lacking the richness of human discourse and often converging toward a narrow set of typical phrasing and structural patterns \cite{jiang2025hivemind}. Recent large-scale studies indicate that modern LLMs produce outputs that are more similar to one another across different model families than human-authored texts \cite{LLMhomogeneity}. This homogenization is characterized by a structural collapse of the probability distribution, often termed mode collapse, where models consistently favor safe, high-probability sequences at the expense of lexical variety \cite{LLMcreativity}. One possible cause of this phenomenon is model alignment or instruction tuning \cite{zhang2026instruction}, implemented using techniques such as Reinforcement Learning with Human Feedback (RLHF) or Direct Preference Optimization (DPO), in which models learn to prioritize familiar-sounding responses that annotators prefer, thereby amplifying distribution collapse \cite{LLMdiversity, FormatCollapse, liu2026alignment}.

A relevant aspect of this homogeneity is how LLMs use vocabulary. Recent studies have shown that text produced by some models has lower lexical diversity \cite{fredrick2025lexical}. A comprehensive evaluation of these features in conversational models reveals that lexical richness is highly sensitive to model parameters, such as presence penalties and temperature, as well as the specific roles assigned to the model \cite{martinez2025beware}. This reduction in variety is particularly evident in the suppression of rare, low-frequency words that exist in the training data but are rarely selected during inference. Interestingly, while human language follows the classical Zipf’s Law \cite{zipf2016human}, frontier LLM outputs across various vendors have been recently found to converge toward a two-parameter Mandelbrot ranking distribution \cite{Mandlebrot} which reveals that LLMs suffer from an artificial steepness in the Zipfian tail, indicating that the probability of selecting rare tokens decays significantly faster than in natural human corpora. This suggests that aligned models are constrained in the depth of the vocabulary they can use.

\begin{figure}[htbp]
  \centering
  \includegraphics[width=0.97\textwidth]{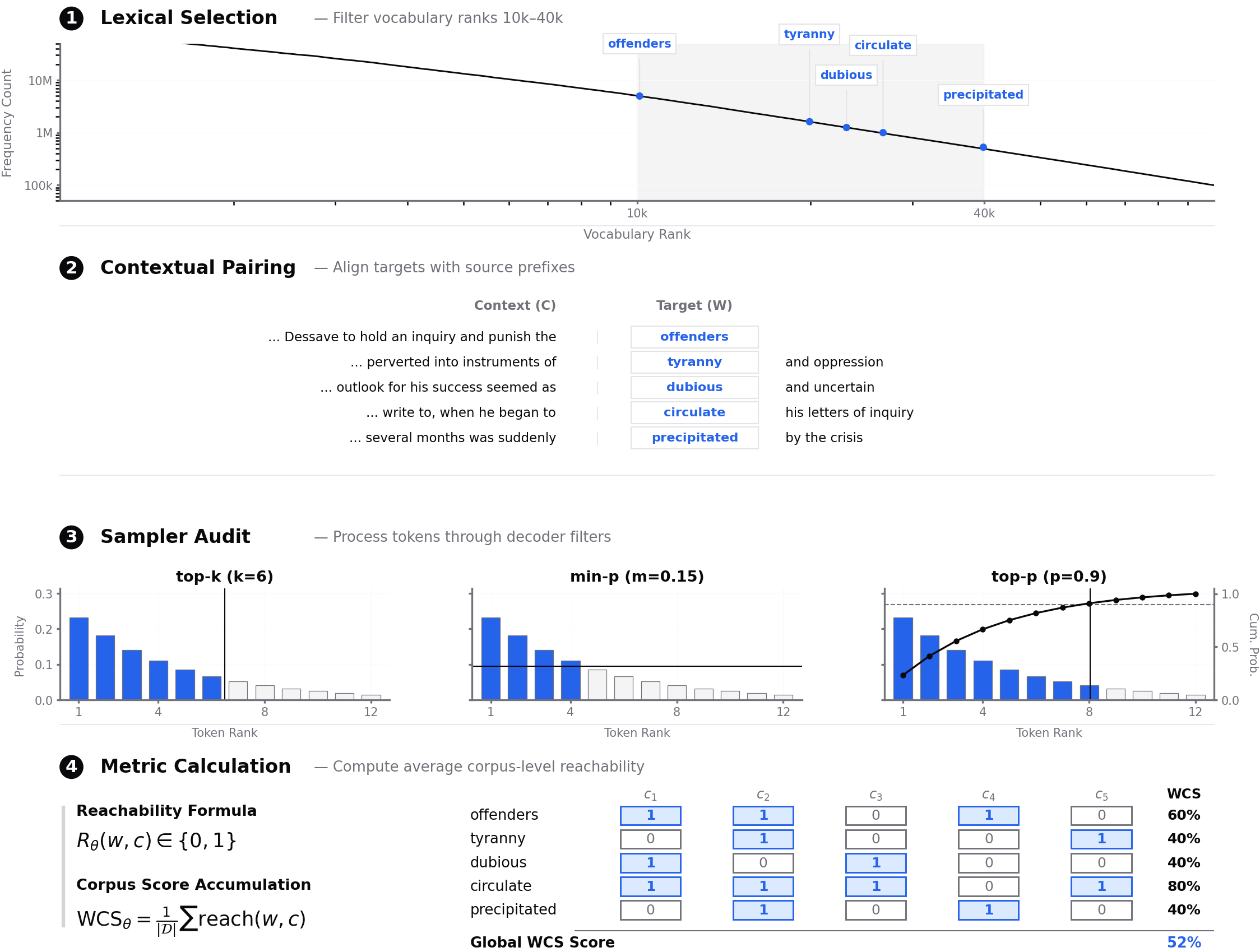}
  \caption{Four-stage methodology for calculating the Word Coverage Score (WCS): select frequency-bounded target words, pair them with human-written contexts, audit token survival under sampling filters, and aggregate reachability into WCS.}
  \label{fig:WCS-diagram}
\end{figure}

Previous research has focused on model knowledge and training data; instead, in this work we investigate the role of decoding mechanics in suppressing linguistic diversity. To quantify this phenomenon, we introduce the \textit{Word Coverage Score} (WCS), a metric that measures the extent to which contextually appropriate human vocabulary is mathematically pruned by standard sampling filters. As illustrated in Figure \ref{fig:WCS-diagram}, the WCS methodology is structured into four core stages designed to isolate the impact of decoding:

\begin{itemize}
    \item \textbf{Stage 1: Lexical Selection:} we identify a target set of ``Middle-Long Tail'' words that represent sophisticated human usage rather than common functional words.
    \item \textbf{Stage 2: Contextual Pairing:}  these words are mapped back into naturalistic human-authored passages to ensure the evaluation occurs within a legitimate linguistic environment.
    \item \textbf{Stage 3: The Forced-Path Audit:}  we perform a \textit{Sampling Audit} to test whether standard filters (e.g., Top-$p$, Top-$k$, or Min-$p$) would mathematically censor that word during generation in that passage.
    \item \textbf{Stage 4: Metric Calculation:} aggregating these results, we identify how the sampler parameters prune the model's latent richness and make it collapse into homogenized output.
\end{itemize}

The rest of the paper is organized as follows. Section \ref{sec:Sampling} examines the process of token selection, framing the sampling process through the metaphor of a "Garden of Forking Paths" and establishing the fundamental trade-off between structural text coherence and vocabulary diversity. Section \ref{sec:Methodology} formally introduces our proposed Word Coverage Score (WCS) framework, detailing the formulation of the Forced-Path Audit alongside our frequency-based lexical and context selection protocols. Section \ref{sec::experiments} outlines the experimental configuration, including the targeted models and samplers used in the evaluation. Section \ref{sec::results} presents our experimental findings, evaluating absolute word erasure and aggregated lexical decay curves across baseline and aligned models. Section \ref{sec::Discussion} provides a high-level discussion on the structural zero-sum constraints of sampling filters, the long-term implications for language evolution, and the inherent limitations of the WCS metric. Finally, Section \ref{sec::Conclusion} concludes the paper.

\section{Token Sampling and LLM Homogeneity}
\label{sec:Sampling}

For each token prediction, LLMs compute the logits for all tokens in their vocabulary, producing tens of thousands of potential candidates. Even if the majority of these candidates possess a negligible probability, the cumulative number of potential trajectories is vast. This configuration creates a literal "Garden of Forking Paths" mirroring Jorge Luis Borges' famous short story, \textit{"El jard{\'i}n de senderos que se bifurcan"} \cite{borges1998garden}. In this narrative, Borges describes an infinite, multi-dimensional labyrinth where every contextually appropriate future step is kept alive simultaneously, allowing parallel timelines to branch out and co-exist. Within an LLM, the presence of sophisticated vocabulary in the latent probability space represents a web of these parallel, co-existing forks. However, the generation of a singular textual sequence requires the decoding system to collapse this multi-dimensional space into a linear path, effectively reverting to what Borges categorizes as traditional fiction where a character selects one alternative and eliminates all the rest.

In practice, the sampling filter acts as an aggressive pruning mechanism that clear-cuts these winding branches even before a final selection is made. Most standard sampling filters, such as Top-$p$, Top-$k$, or Min-$p$, initially truncate the vocabulary distribution to eliminate low-probability forks, forcing the final token selection to occur exclusively among the surviving candidates. The primary, explicit objective of this truncation is to prune the highly volatile long-tail of the predictive distribution, thereby trapping the model within high-probability paths to ensure semantic consistency, structural grammar, and local coherence \cite{holtzman2019curious}. We argue that by doing so, these standard decoders systematically restrict the model's expressive aperture, resulting in a flat, uniform discourse stripped of the rich human expression originally envisioned in Borges' garden. This severe restriction artificially constrains the operational vocabulary used by LLMs, effectively erasing viable words from the generative landscape as quantified by our Word Coverage Score (WCS).

This truncation exposes a fundamental dilemma inherent to the autoregressive generation paradigm: a practical trade-off between local text coherence and global linguistic diversity. To reduce the risk of low-quality or less coherent generation when sampling from unconstrained, long-tail distributions, modern token sampling filters explicitly prioritize high-probability predictability at the cost of reducing lexical diversity. Because current architectures rely heavily on token-by-token selection from bounded distributions to maintain sequential continuity, they may struggle to simultaneously protect the sophisticated nuances of human vocabulary. Consequently, this architectural constraint imposes an artificial upper bound on the expressiveness and vocabulary of modern LLMs, presenting an intrinsic limitation of current autoregressive language generation.

\section{Methodology}
\label{sec:Methodology}

The Word Coverage Score (WCS) quantifies the differences between a model's latent lexical knowledge and its generative accessibility under specific decoding constraints. Rather than evaluating an LLM's static capability to comprehend a word in isolation, the WCS operates as a dynamic behavioral metric that measures whether a contextually appropriate, human-authored vocabulary choice remains mathematically reachable during sequential autoregressive generation.  By formalizing reachability as a product of step-wise token survival, the WCS provides a rigorous framework to measure how token sampling reduces the vocabulary used. The following subsections describe each of the elements of the proposed WCS.

\subsection{The Forced-Path Audit}

To evaluate the WCS, we employ a \textit{Forced-Path Audit}. Given a human-authored reference sequence $S$ from an evaluation corpus $\mathcal{X}$, we identify a target lexical unit or word $w$ composed of $n$ sub-word tokens $(t_1, t_2, \dots, t_n)$. We provide the model with the ground-truth prefix context $C$ and force a deterministic traversal of the path $w$.

At each transition $i$, we extract the full probability distribution $P(\cdot \mid C, t_1, \dots, t_{i-1})$. Rather than sampling, we record the rank and scalar probability of the ground-truth token $t_i$. This allows us to determine if $t_i$ would have survived the pruning logic of a given sampling algorithm.

\subsection{Lexical Survival Functions}

We define Reachability ($\mathcal{R}$) as a binary indicator of whether a token remains in the "active" vocabulary set after a sampling filter is applied. A multi-token word $w$ is considered covered ($\mathcal{R}=1$) if, and only if, every constituent token $t_i$ survives the filter at its respective step. In our evaluation we focus on traditional sampling techniques such as top-$k$ and top-$p$ but we also include Min-$p$ as a representative of emerging sampling techniques that try to balance text coherence and diversity \cite{minp, ding2026mink, tang2024topnsigma}. In more detail, we define three primary survival functions for a word $w$ and a context $C$ corresponding to those decoders:

\begin{itemize}
    \item \textbf{Top-$k$ Survival:} The token must rank within the $k$ most probable outcomes.
    \begin{equation}
        \mathcal{R}_k(w,c) = \prod_{i=1}^{n} \mathbb{I}(\text{rank}(t_i) \leq k)
    \end{equation}

    \item \textbf{Top-$p$ (Nucleus) Survival:} The token must belong to the smallest set $V^{(p)}$ whose cumulative probability meets threshold $p$.
    \begin{equation}
        \mathcal{R}_p(w,c) = \prod_{i=1}^{n} \mathbb{I}(t_i \in V^{(p)}) \quad \text{where} \sum_{v \in V^{(p)}} P(v) \geq p
    \end{equation}

    \item \textbf{Min-$p$ Survival:} The token's probability must exceed a scaled fraction $m$ of the maximum token probability $P(t_{top})$.
    \begin{equation}
        \mathcal{R}_m(w,c) = \prod_{i=1}^{n} \mathbb{I}\left(P(t_i) \geq m \cdot P(t_{top})\right)
    \end{equation}
\end{itemize}

The inclusion of Min-$p$ allows for an assessment of its capacity to preserve the diversity of the model's distribution compared to more traditional, rank-based pruning methods.

\subsection{The Word Coverage Score (WCS)}

The Word Coverage Score is the aggregate mean reachability across a target set of words $W$, each with a set of contexts $C_w$ under a specific parameter configuration $\theta \in \{k, p, m\}$.

\begin{equation}
WCS(\theta) = \frac{1}{\sum_{w \in W} |C_w|} \sum_{w \in W} \sum_{c \in C_w} \mathcal{R}_\theta(w, c)
\end{equation}

By calculating $WCS(\theta)$ across a continuous range of values, we generate a lexical decay curve. This function allows for the identification of how the sampler parameters impact the potential lexical richness.

The use of several contexts per word enables us to also do a per-word calculation of the WCS as follows:

\begin{equation}
WCS(\theta, w) = \frac{1}{|C_w|} \sum_{c \in C_w} \mathcal{R}_\theta(w, c)
\end{equation}

which is of interest to analyze the reachability per word. For example, when a word is not reachable for any of its context we can argue that it has been "removed" by the sampler.

\subsection{Frequency-Based Lexical Selection}

To ensure a statistically representative and unbiased evaluation of lexical reachability, we utilize a Band-Limited Random Sampling protocol focusing on words that are not common but not extremely rare. We establish our target lexical set $W$ using the Google Web Trillion Word Corpus as compiled by \cite{norvig2009}. This source was selected specifically because its frequency distribution closely mirrors the large-scale web-crawled datasets (e.g., Common Crawl) typically used in the pre-training of Large Language Models (LLMs). From the total corpus, we isolate the "Middle-Long Tail" band by selecting words ranked between $10,000$ and $40,000$ in total frequency. This band was chosen specifically to avoid high-frequency functional words (ranks $<10,000$) where reachability should not be an issue, and extremely low-frequency noise (ranks $>40,000$) where model training data may be insufficient.

The words are selected randomly from that range, and before adding them to the lexical set $W$, we apply a Dictionary-Validation Filter. Each candidate word was cross-referenced against the Moby Word Lists \cite{ward2002moby} to remove non-lexical artifacts such as URLs, OCR errors, and technical metadata while preserving the authentic frequency-based sampling of the web-scale corpus.

\subsection{Context Selection}

To evaluate lexical reachability in a complex, long-context environment, we utilize the PG-19 dataset \cite{rae2019compressive}. Derived from the Project Gutenberg library, PG-19 contains full-length books published before 1919, providing a linguistically rich and diverse vocabulary that reflects human diversity and avoids the stylistic homogenization common in modern web-crawled datasets.

Each word in $W$ is then mapped back to $C$ naturalistic contexts within the \textit{PG-19} corpus for evaluation, ensuring that the WCS measures the reachability of legitimate human expressions. The following procedure is used to generate each of the $C$ contexts:

\begin{enumerate}
    \item \textbf{Random Initialization:} For each target word $w$, we select a random byte-offset within the \textit{PG-19} test partition and perform a forward linear search for the first occurrence of $w$.
    \item \textbf{Contextual Extraction:} Upon identification, we extract the preceding $L=256$ tokens to serve as the prefix $C$. If the word appears within the first 256 tokens of a document, we continue the search to the next occurrence to ensure a full contextual window.
    \item \textbf{Context Verification:}To ensure semantic integrity, each extracted context was evaluated for coherence using the \texttt{gemini-2.5-flash} model \cite{gemini2025flash} in a zero-shot binary classification setup. Contexts classified as non-coherent, containing artifacts such as table of contents or index fragments, or lacking sufficient linguistic information are discarded and replaced via the sampling protocol.
\end{enumerate}

By using a random-entry search rather than a top-down approach, we ensure that our contextual samples are distributed across the entire breadth of the corpus, capturing a diverse range of narrative styles and positions within the source documents.

\section{Experiments}
\label{sec::experiments}

The experimental framework is designed to empirically trace how a word navigates from a model's latent internal distribution to its final generation sequence. To achieve this, our methodology is segmented into three logical phases: first, we establish a list of foundational and aligned model architectures; second, we formalize a rigorous tracking protocol that audits the step-by-step token survivability of target human vocabulary; and third, we systematically evaluate word reachability under common truncation thresholds and temperature scales. The specific implementations of these experimental stages are detailed below.

\subsection{Model Selection}

To isolate the impact of alignment and distillation on lexical diversity, we evaluate both the Base (raw) and Instruct/It (aligned) variants of leading open-weight architectures with fewer than 20 billion parameters. This selection shown in Table \ref{tab:models} ensures a broad cross-section of the current LLM landscape while maintaining computational accessibility for independent researchers. For each architecture, the Base version represents the model's fundamental linguistic distribution, while the Instruct/it version represents the distribution post-alignment (RLHF/SFT). For models like DeepSeek-R1-Distill-Qwen-14B, which do not have a "native" base in the traditional sense, we use the corresponding Qwen-14B-Base as the reference point. This allows us to measure the specific "Distillation Deficit", that is the lexical loss incurred when a base distribution is forced to mirror the reasoning chains of a larger teacher model.

\begin{table}[h]
\centering
\begin{tabular}{l l l}
\hline
\textbf{Family} & \textbf{Base Variant} & \textbf{Aligned Variant} \\
\hline
Llama & Llama-3.1-8B~\cite{hf-llama31-8b} & Llama-3.1-8B-Instruct~\cite{hf-llama31-8b-instruct} \\
Mistral & Mistral-7B-v0.3~\cite{hf-mistral7b-v03} & Mistral-7B-Instruct-v0.3~\cite{hf-mistral7b-instruct-v03} \\
Qwen (Small) & Qwen3.5-9B-Base~\cite{hf-qwen35-9b-base} & Qwen3.5-9B~\cite{hf-qwen35-9b} \\
Qwen (Mid) & Qwen2.5-14B~\cite{hf-qwen25-14b} & Qwen2.5-14B-Instruct~\cite{hf-qwen25-14b-instruct} \\
Gemma-3 & Gemma-3-12B-pt~\cite{hf-gemma3-12b-pt} & Gemma-3-12B-it~\cite{hf-gemma3-12b-it} \\
Gemma-4 & Gemma-4-E4B~\cite{hf-gemma4-e4b} & Gemma-4-E4B-it~\cite{hf-gemma4-e4b-it} \\
DeepSeek & Qwen2.5-14B~\cite{hf-qwen25-14b} & DeepSeek-R1-Distill-Qwen-14B~\cite{hf-deepseek-r1-qwen14b} \\
\hline
\end{tabular}
\caption{Selected model pairs for evaluating the Word Coverage Score (WCS).}
\label{tab:models}
\end{table}

\subsection{Audit Protocol}

For each model pair, we select $N_w=100$ words and for each of them $N_c=10$ contexts. We define a trial as a "success" (Reachability $\mathcal{R}=1$) only if the complete multi-token sequence of the target word $w$ remains within the valid sampling set $\mathcal{V}_\theta$ for a given parameter $\theta$.

\subsection{Sampling Sweep}

We systematically sweep the sampling decoders to identify the impact on word coverage:

\begin{enumerate}
    \item \textbf{Nucleus ($p$):} $[0.7, 0.8, 0.9, 0.95, 0.99]$ in increments of $0.05$.
    \item \textbf{Top-$k$:} $[1, 20]$ in increments of $1$.
    \item \textbf{Min-$p$:} $[0.01, 0.1]$, evaluating its capacity to preserve the diversity of the distribution compared to rank-based pruning.
\end{enumerate}

Additionally, we conduct experiments for three different settings $T = 0.7,1,1.5$. The first two values correspond to settings commonly used for chat or text writing applications, while $T=1.5$ is included as an aggressive high-temperature condition that is less commonly used in practice. Prior work on narrative generation finds that increasing temperature is only weakly associated with novelty and is also correlated with reduced coherence, suggesting a trade-off rather than a simple creativity control \cite{peeperkorn2024temperature}.

\section{Results}
\label{sec::results}

The code and results are available at \url{https://github.com/WordsGPT/WCS} to facilitate further analysis and reproducibility. The repository also includes a static interactive visualiser at \url{https://wordsgpt.github.io/WCS/temperature.html} for the results, allowing the WCS curves to be inspected across models, samplers, parameters, temperatures, and aggregation levels. We present our results by plotting the $WCS$ for each model versus the sampler parameters. First, we look at the word level, analyzing the percentage of words that are not reachable in any of their $N_c=10$ contexts, i.e., $WCS(\theta,w)=0$. Note that these are words that the model will never select under the sampling algorithm and configuration parameters in the ten text contexts, so the words are effectively "erased".

\begin{figure}[h]
    \centering
    \includegraphics[width=0.55\textwidth]{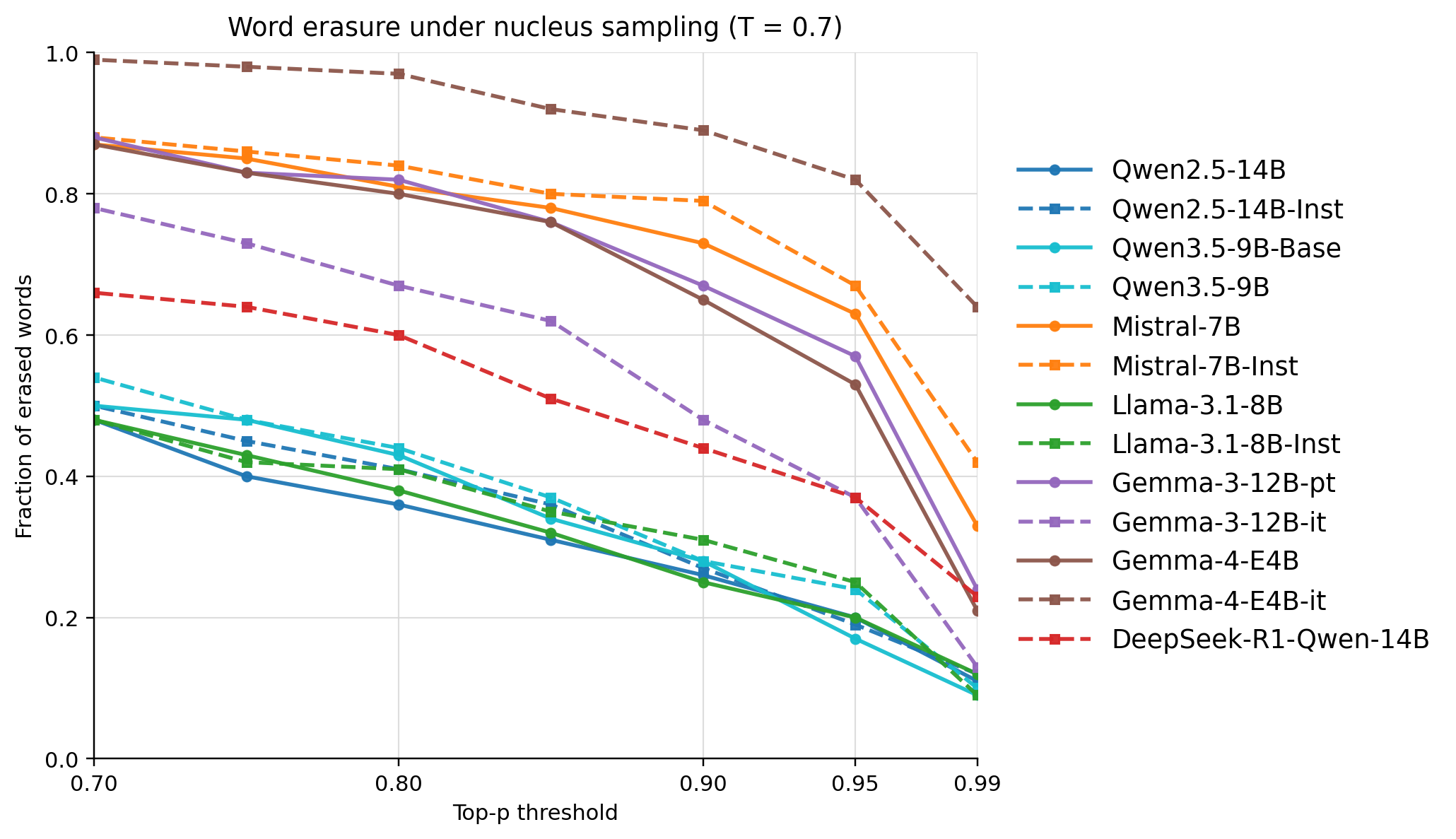}
    \caption{Fraction of words with $WCS(\theta,w)=0$ when using Nucleus Sampling ($p$) and $T=0.7$. Instruct models (dashed lines) and their Base counterparts (solid lines). Default and recommended multi-parameter sampler settings are reported separately in Table \ref{tab:default-sampling-settings}.}
    \label{fig:TopP-T07}
\end{figure}

The results for Top-$p$ sampling with a temperature of $T=0.7$ are shown in Figure \ref{fig:TopP-T07}. It can be observed that even for $p = 0.95$, most of the models have a significant fraction of words that cannot be sampled in any of the contexts, showing that lexical reachability is poor for the words evaluated. In fact, even increasing to $p = 0.99$ still leaves many words out of the sampling. When comparing base models (solid lines) and their instruction-tuned (instruct/it) counterparts (dashed lines), a consistent trend emerges across most families: with the exception of Gemma-3-12B, the aligned versions exhibit a larger fraction of erased words relative to their baseline variants. This provides strong evidence that preference optimization processes (such as RLHF, or DPO) have in most cases a direct, restrictive impact on generative diversity in terms of the usable vocabulary space.

Across models, the Gemma family has non-uniform patterns between generations. While the pre-trained Gemma-3-12B-pt and Gemma-4-E4B demonstrate comparable baseline results, alignment yields opposite effects: it reduces word erosion in Gemma-3-12B-it, whereas it exacerbates it in Gemma-4-E4B-it, effectively eliminating the majority of the evaluated words from the reachable distribution. In contrast, the Qwen and Llama families demonstrate relatively lower overall absolute word erosion, with their respective instruct versions tracking closely to their base counterparts, but reducing vocabulary diversity. Finally, DeepSeek-R1-Distill-Qwen-14B exhibits a significant vocabulary loss compared to the underlying Qwen2.5-14B base model from which it is derived. This pattern suggests that the distillation of structured reasoning capabilities introduces an additional constraint on lexical accessibility.

Averaging across matched base/instruct model pairs at the same temperature, sampler, and parameter settings, aligned models exhibit a small but consistent reduction in lexical reachability: mean word-level reachability decreases from $0.740$ for base models to $0.728$ for instruct models, while WCS decreases from $0.290$ to $0.277$. This aggregate trend is not universal, with Gemma-3-12B-it improving over its pre-trained counterpart, but most matched families show lower reachability after instruction tuning.

\begin{figure}[h]
    \centering
    \includegraphics[width=0.55\textwidth]{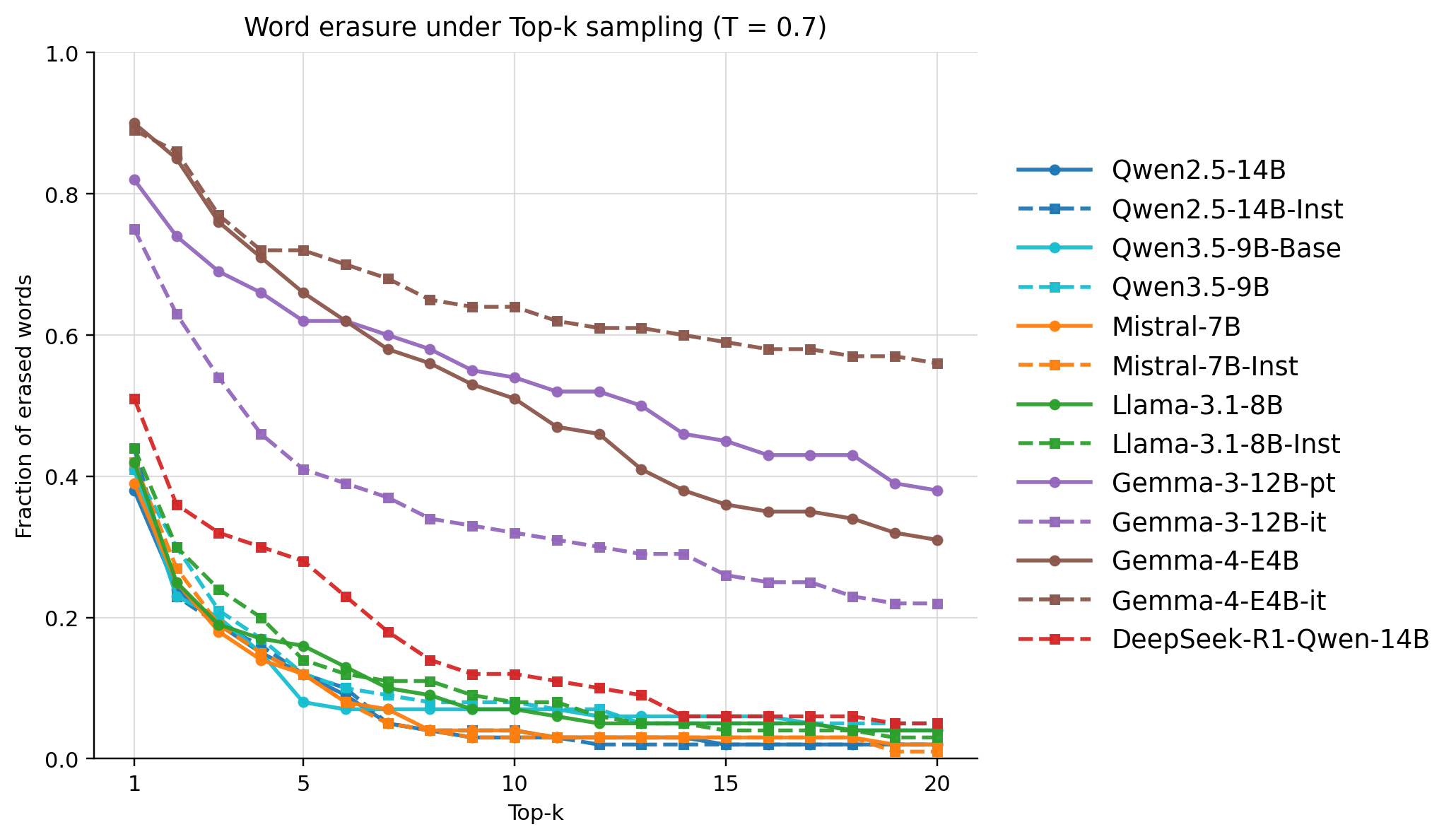}
    \caption{Fraction of words with $WCS(\theta,w)=0$ when using Top-$k$ Sampling and $T=0.7$. Instruct models (dashed lines) and their Base counterparts (solid lines).}
    \label{fig:TopK-T07}
\end{figure}

\begin{figure}[h]
    \centering
    \includegraphics[width=0.55\textwidth]{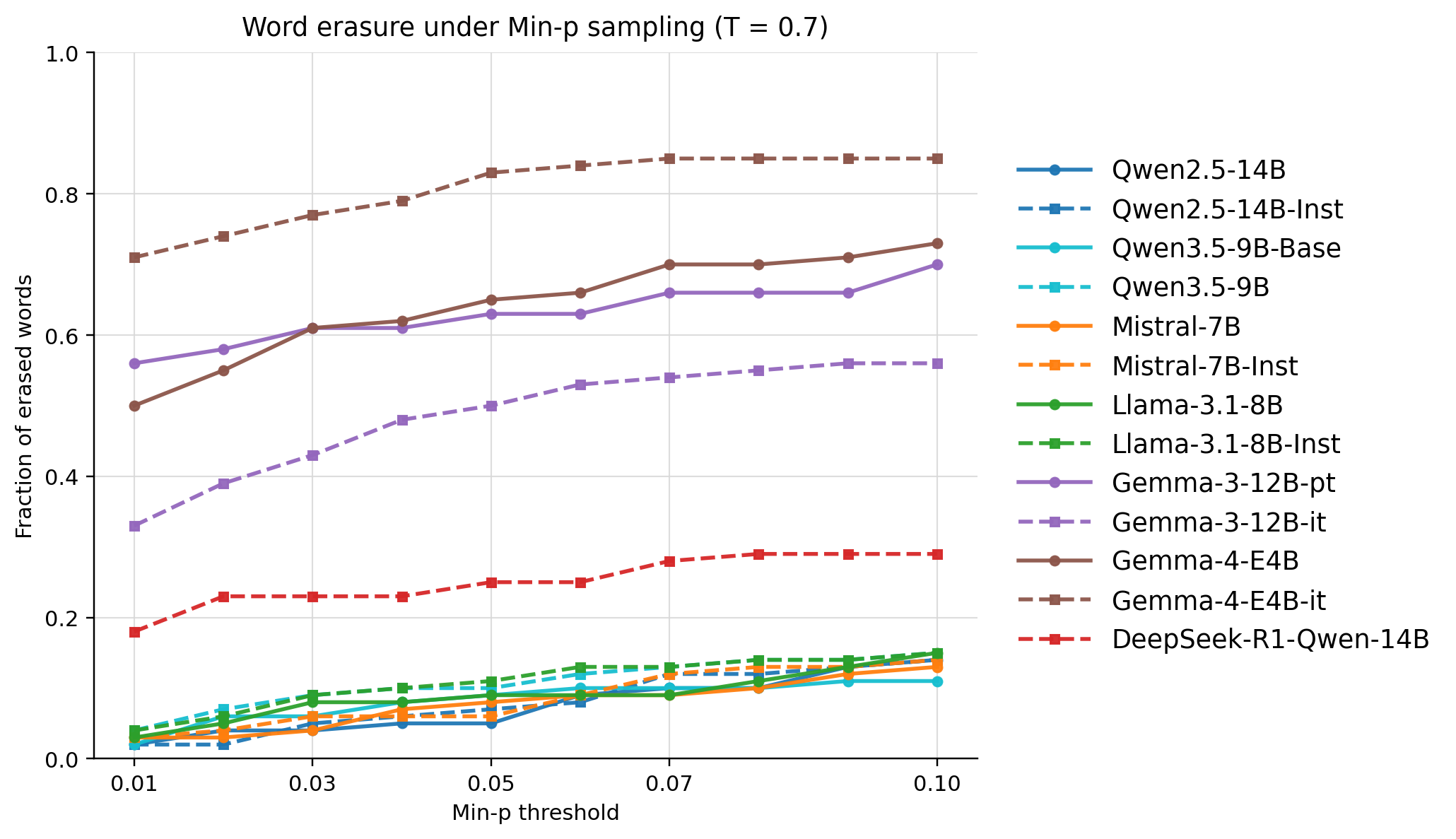}
    \caption{Fraction of words with $WCS(\theta,w)=0$ when using Min-$p$ Sampling and $T=0.7$. Instruct models (dashed lines) and their Base counterparts (solid lines).}
    \label{fig:minp-T07}
\end{figure}

\clearpage

Nucleus sampling with values of $p$ in the range of 0.8 to 0.95 and temperature at or below 1.0 is commonly used in model cards and generation configurations. Therefore, it is of interest to analyze word erosion under these commonly used settings to get an idea of the impact of current samplers and settings on word reachability. For the models evaluated, the documented settings include Qwen2.5-14B-Instruct with Top-$p=0.8$, Top-$k=20$, and $T=0.7$ in its Hugging Face generation configuration\footnote{\url{https://huggingface.co/Qwen/Qwen2.5-14B-Instruct/blob/main/generation_config.json}}; Qwen3.5-9B with recommended non-thinking general-purpose settings of Top-$p=0.8$, Top-$k=20$, and $T=0.7$, and thinking-mode settings of Top-$p=0.95$, Top-$k=20$, and $T=1.0$\footnote{\url{https://huggingface.co/Qwen/Qwen3.5-9B}}; Llama-3.1-8B-Instruct with a generation configuration of Top-$p=0.9$ and $T=0.6$\footnote{\url{https://huggingface.co/meta-llama/Llama-3.1-8B-Instruct/blob/main/generation_config.json}}; Gemma-4-E4B with Top-$p=0.95$, Top-$k=64$, and $T=1.0$\footnote{\url{https://huggingface.co/google/gemma-4-E4B/blob/main/generation_config.json}}; and DeepSeek-R1 models with a recommended temperature range of $0.5$--$0.7$ ($0.6$ recommended) and benchmark sampling at Top-$p=0.95$\footnote{\url{https://huggingface.co/deepseek-ai/DeepSeek-R1-Distill-Qwen-14B}}. Mistral-7B-Instruct-v0.3 exposes the same sampling controls in common inference stacks, but its model card does not specify a single default Top-$p$ value\footnote{\url{https://huggingface.co/mistralai/Mistral-7B-Instruct-v0.3}}.

Table \ref{tab:default-sampling-settings} reports word-level reachability and WCS under documented default or recommended decoding settings for representative evaluated models. These values complement the one-dimensional sampler sweeps: Figure \ref{fig:TopP-T07} isolates the effect of varying Top-$p$, whereas practical decoding configurations may combine Top-$p$, Top-$k$, and temperature. Under these commonly used settings, word erosion is severe with 22\% to 57\% of words not being reachable in any of the contexts and only between 59 and 230 of the 1,000 contexts being reachable. The worst performance is for the Gemma-4-E4B-it model whose default configuration covers at least one context for only $43\%$ of target words, and only $5.9\%$ of all evaluated contexts. These results illustrate that word erosion is not an anomaly, but a systemic characteristic of current inference systems, where standard sampling parameters suppress lexical diversity.

\begin{table}[h]
\centering
\resizebox{\textwidth}{!}{%
\begin{tabular}{l l c c c}
\hline
\textbf{Model / family} & \textbf{Default setting} & \textbf{Words reachable} & \textbf{Erased words} & \textbf{WCS} \\
\hline
Qwen2.5-14B-Instruct & $p=0.8$, $k=20$, $T=0.7$ & $0.57$ & $43\%$ & $0.144$ \\
Qwen3.5-9B & $p=0.8$, $k=20$, $T=0.7$ & $0.52$ & $48\%$ & $0.124$ \\
Llama-3.1-8B-Instruct & $p=0.9$, $T=0.7$ & $0.69$ & $31\%$ & $0.230$ \\
Gemma-4-E4B & $p=0.95$, $k=64$, $T=1.0$ & $0.78$ & $22\%$ & $0.173$ \\
Gemma-4-E4B-it & $p=0.95$, $k=64$, $T=1.0$ & $0.43$ & $57\%$ & $0.059$ \\
DeepSeek-R1-Distill-Qwen-14B & $p=0.95$, $T=0.7$ & $0.63$ & $37\%$ & $0.160$ \\
\hline
\end{tabular}%
}
\caption{WCS under documented default or recommended decoding settings for representative evaluated models. Words reachable is the fraction of target words reachable in at least one of their contexts; erased words are those with $WCS(\theta,w)=0$. WCS is the aggregate $WCS(\theta)$ over all evaluated word-context pairs. Rows with both Top-$p$ and Top-$k$ use the combined filter. For Llama and DeepSeek, $T=0.7$ is the nearest evaluated temperature to the documented recommendations. Mistral-7B-Instruct-v0.3 is omitted because its model card does not specify a single default Top-$p$ value.}
\label{tab:default-sampling-settings}
\end{table}



The results for Top-$k$ and Min-$p$ are shown in Figures \ref{fig:TopK-T07},\ref{fig:minp-T07}. Word coverage can be seen to improve for most models, but there are still models that lose a significant fraction of words even under aggressive settings. The Gemma family shows again a different pattern across generation: Gemma-3-12B-pt and Gemma-4-E4B behave broadly like the other base models, while Gemma-4-E4B-it retains substantially lower word reachability under Top-$k$ and Min-$p$ than its base counterpart. This gap is also clear under Gemma-4's default settings, Top-$p=0.95$, Top-$k=64$, and $T=1.0$, as shown in Table \ref{tab:default-sampling-settings}. Gemma-4-E4B has a WCS of $0.173$, whereas Gemma-4-E4B-it falls to $0.059$. Since the Gemma models have the same tokenizer, the differences can only be attributed to training (Gemma-3 vs Gemma-4) or alignment (base versus instruction tuned), showing the impact of these procedures in preserving words.


\begin{figure}[h]
    \centering
    \includegraphics[width=0.6\textwidth]{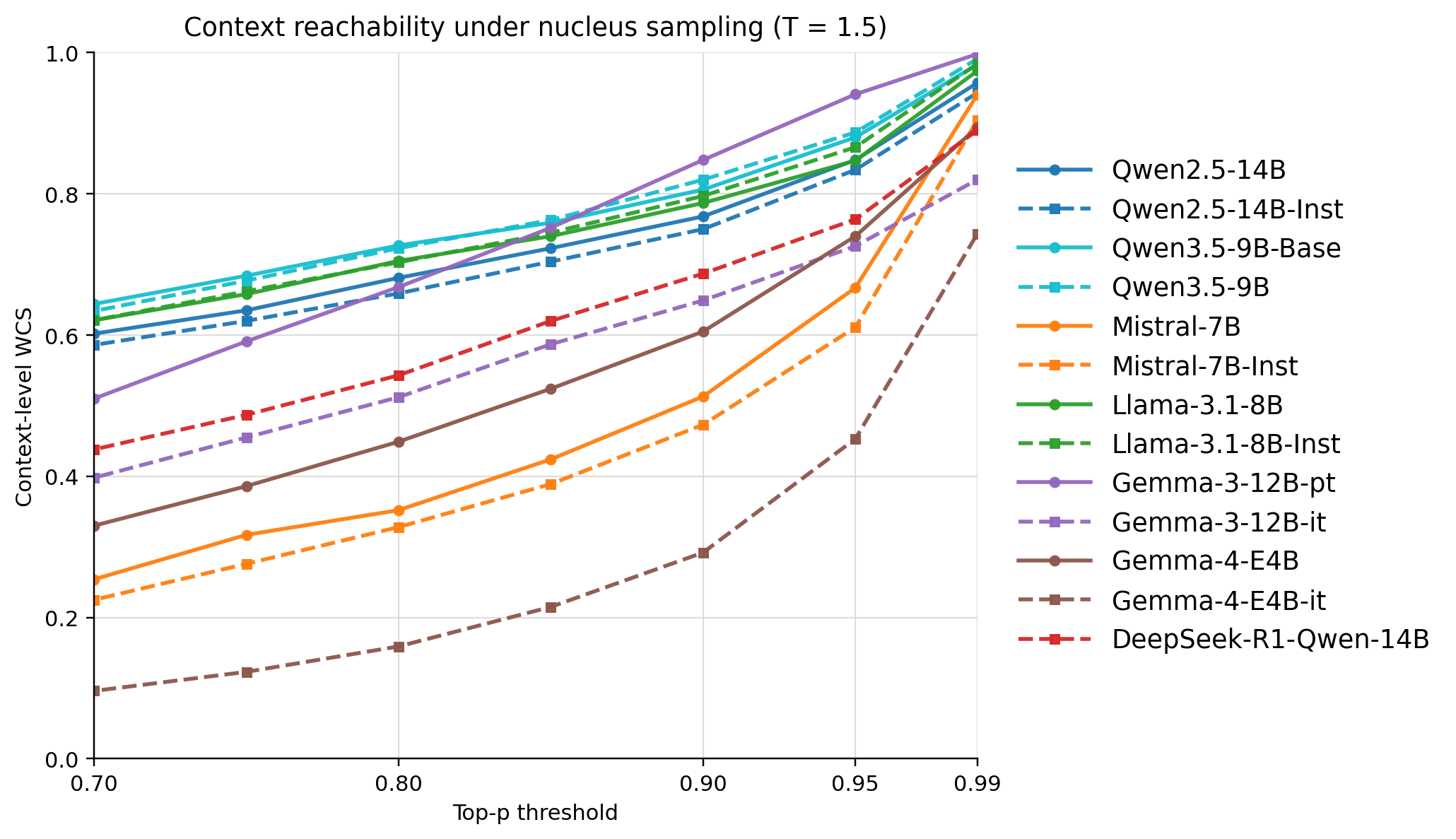}
    \includegraphics[width=0.6\textwidth]{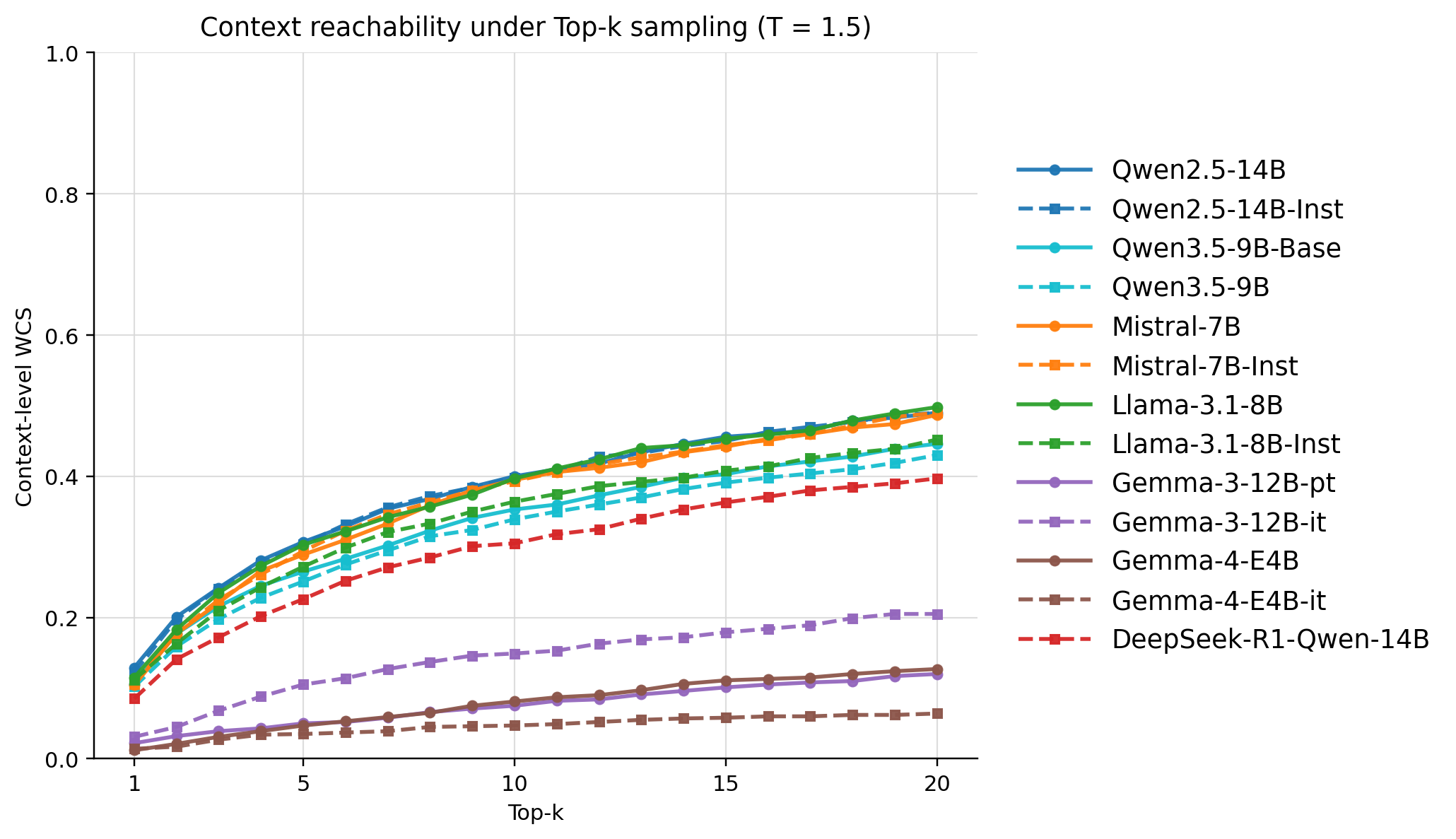}
    \includegraphics[width=0.6\textwidth]{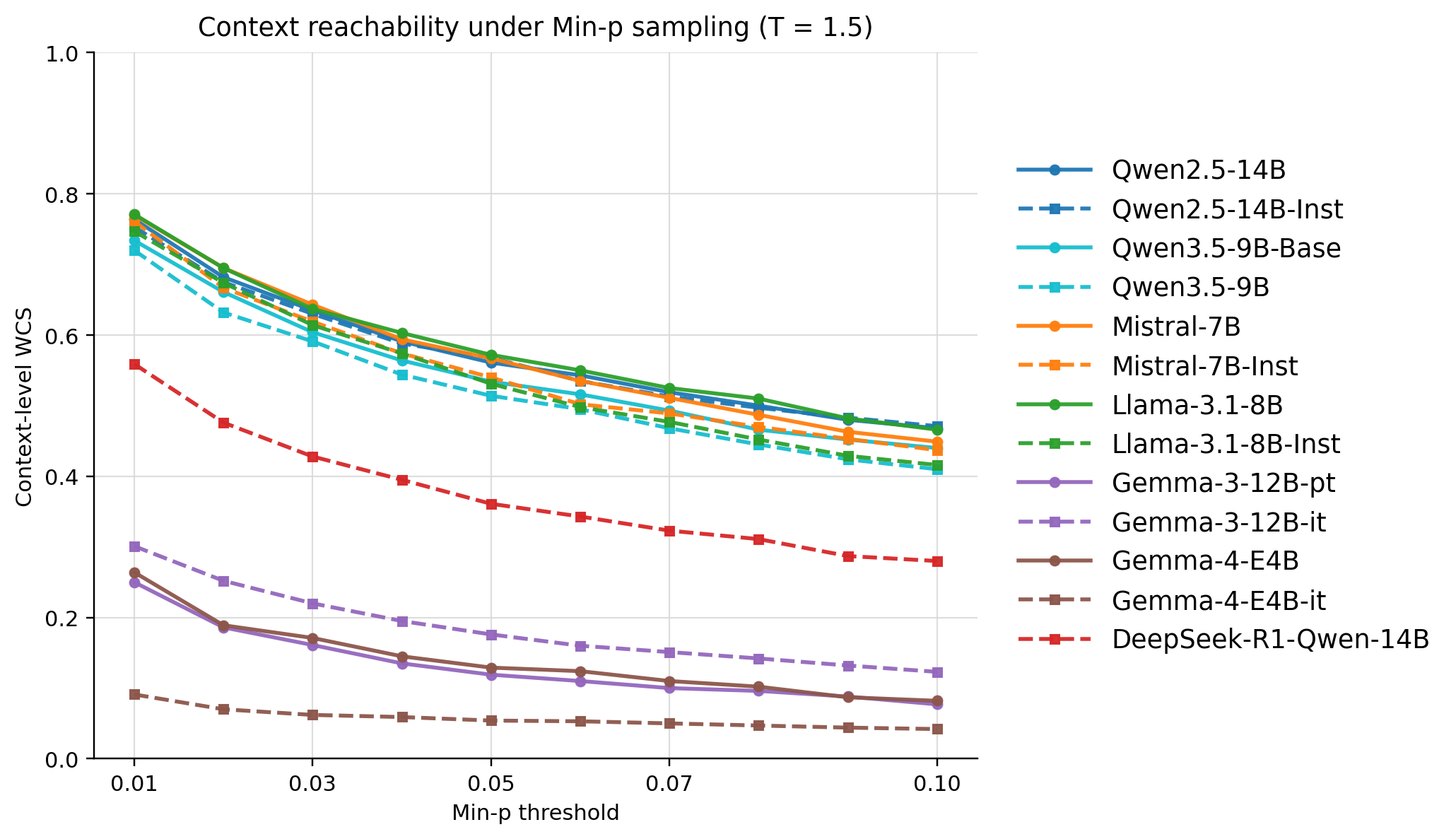}
    \caption{Fraction of the 1,000 contexts (100 words each with 10 contexts) that are reachable  $WCS(\theta)$ when using Top-$p$ (top), Top-$k$ (top), Min-$p$ (bottom) Sampling and $T=1.5$. Instruct models (dashed lines) and their Base counterparts (solid lines).}
    \label{fig:WCS}
\end{figure}

In the previous analysis we have focused on the extreme case of words for with $WCS(\theta,w)=0$, i.e. they will never be sampled by the models in any of the ten contexts selected from the corpus. Those words are completely removed; however, it is also interesting to look at the effects of sampling at a higher aggregate level. To do so, in Figure \ref{fig:WCS} we show the percentage of contexts for all the words that can be selected by each sampling algorithm with $T= 1.5$. It can be clearly observed that even under this aggressive high-temperature setting, a significant fraction of the contexts will never select the word. This implies that even when words are not completely removed from all their 10 contexts, they suffer a significant erosion that reduces their use in texts where they should be one of the options considered for completion. This illustrates that even when not completely "removing" a word, models can limit their use which is another form of lexical degradation and diversity reduction.


\section{Discussion}
\label{sec::Discussion}

In this section, the results are discussed at a higher level of abstraction, focusing on their implications for LLM text generation and their limitations.

\subsection{Interpretation of the Word Coverage Score}

When analyzing our experimental findings, it is important to discuss the information captured by the Word Coverage Score (WCS). The WCS fundamentally operates as a metric of binary exclusion: it isolates an extreme manifestation of word erosion by detecting instances where a target word becomes mathematically impossible ($\mathcal{R}=0$) to generate under a given context and sampling configuration. However, lexical erosion in natural language generation is not an all-or-nothing phenomenon; it can also manifest as a probability dampening.

In less severe scenarios, a sampling filter may allow a sophisticated word to remain within the truncated candidate pool ($\mathcal{V}_\theta$), yet reduce its assigned probability mass significantly below the threshold observed in natural human discourse. Such a word remains technically reachable, but its generational frequency is reduced, leading to a subtler form of soft erosion. Because the WCS focuses strictly on the total elimination of linguistic paths it does not account for this soft probabilistic erosion. Consequently, our empirical results must be interpreted as a conservative lower bound on the true scale of word erosion; the absolute loss of human lexical diversity in practical applications is likely larger due to these cumulative, non-binary suppression effects that escape the WCS metric.

\subsection{The Zero-Sum Game of Sampling}

Our empirical results suggest a conflicting relationship between lexical reachability and structural text coherence. To recover sophisticated, human-utilized words lost in the $10\text{k}\text{--}40\text{k}$ frequency band, generative configurations often require elevating $p$ near unity or increasing the temperature beyond typical defaults. However, this is not a complete solution: even at $T=1.5$, our results show that many words remain unreachable in some contexts. This aligns with evidence that higher temperature can modestly increase novelty while correlating with reduced coherence \cite{peeperkorn2024temperature}. Consequently, practical decoding involves a trade-off: less restrictive settings can improve lexical reachability, while more conservative settings better preserve coherence but can also remove viable human vocabulary from the candidate set.

\subsection{Implications for Language Evolution}

The systemic use of restrictive sampling mechanics to enforce structural coherence inevitably induces a form of lexical erosion: an artificial and permanent constriction of the active human lexicon to a compressed subset of high-probability vocabulary. This structural narrowing has profound implications for the broader landscape of  language evolution. As model-generated text increasingly populates public repositories, educational frameworks, and digital publications, future generations of language models will recursively ingest this flattened discourse as training data. Consequently, the rich, expressive "Long Tail" historically characterized by the Zipfian and Norvig distributions \cite{zipf2016human}, \cite{norvig2009} risks being erased. By replacing the diverse, exploratory, and heterogeneous textures of human literacy with automated, homogenized text, standard inference decoders do not merely prune tokens in the short term, they may actively reshape future human language evolution.

\subsection{Mitigating Word Erosion}

The most direct path to reduce word erosion is to explore alternative sampling schemes and also to optimization of the sampler hyperparameters. However, we believe that mitigating word erosion demands a more fundamental shift in decoding pipelines. Current models treat sampling as a layer completely decoupled from the model's underlying semantic understanding. Future research could explore semantic-guided decoders that can actively differentiate between low-probability data noise from contextually rich, low-frequency human vocabulary. By integrating external lexical constraints directly within the inference loop, models can safely expand their expressive capacity.

Another possible direction is to modify the training objectives of Large Language Models by incorporating explicit lexical diversity and reachability penalties directly into the loss function. For example, to directly reward the maintenance of an expansive "Garden of Forking Paths" across the $10\text{k}\text{--}40\text{k}$ Zipfian band. Furthermore, preference optimization frameworks could be modified to reward stylistic and lexical variation rather than purely optimizing for flat, conversational uniformity. By treating vocabulary reachability as an optimization metric, models may be able to better preserve the vocabulary.

Ultimately, reversing the trend toward digital linguistic homogenization requires treating vocabulary diversity not as a superficial variable, but as a critical architectural objective equivalent to factual correctness and syntactic coherence.

\section{Conclusion}
\label{sec::Conclusion}

In this work, we have investigated the role of token sampling in the linguistic homogenization text generated by Large Language Models. We introduce the Word Coverage Score (WCS) which provides a rigorous framework to isolate the direct impact of inference-time token truncation in vocabulary use. By auditing step-by-step token survivability through a naturalistic, long-context Forced-Path Audit, the WCS exposes a divergence between a model's internal latent vocabulary knowledge and its empirical generative reachability under standard operational constraints.

Our empirical results across several open-weight models demonstrate that typical decoding filters act as pruning mechanisms that fundamentally restrict a model's vocabulary use. In the contextually rich $10\text{k}\text{--}40\text{k}$ word frequency band, samplers with default parameters systematically erase meaningful lexical paths and do not use a significant fraction of words in that frequency range. The implications of this finding pose a profound challenge to language evolution. As automated text recursively populates public data repositories, future iterations of language models will inevitably ingest this flattened discourse, creating a loop that will most likely lead to more homogeneous language.

Mitigating word erosion requires a fundamental paradigm shift: first, toward semantic-guided decoders capable of distinguishing contextually rich expression from background token noise, and second, toward re-engineering training and preference optimization objectives to treat lexical breadth as a first-class loss penalty. Treating vocabulary diversity not as a superficial parameter, but as a core architectural objective equivalent to factual correctness and syntactic grammar is imperative to keep the infinite, parallel branches of Borges' garden alive.

\section{Limitations}
\label{sec:limitations}

Although the Word Coverage Score (WCS) provides a general method for detecting and quantifying lexical erosion inference-time, several limitations must be acknowledged to better frame our results and guide future extensions of this work.

\paragraph{Model Architecture and Parameter Constraints:}
Our empirical evaluation is strictly confined to open-weight model architectures with fewer than 20 billion parameters. This threshold was selected deliberately to ensure computational accessibility and reproducibility for independent researchers. However, it leaves the behaviors of large frontier models unexplored. Furthermore, because our methodology relies fundamentally on extracting complete, step-wise logit probability distributions at each token transition, this protocol cannot be extended to proprietary, closed-source models whose API endpoints only yield compiled text outputs or highly restricted top-$k$ logit values.

\paragraph{Linguistic and Historical Bias:}
The entire analytical pipeline was optimized for and restricted to the English language. The underlying corpora utilized for frequency estimation and verification are  English-centric. Additionally, our context selection is derived from the PG-19 dataset, which consists exclusively of human-authored books published prior to 1919. This choice circumvents the systemic stylistic homogenization seen in modern web-scraped text data but introduces a historical bias.

\paragraph{Boundary Bounds of Frequency Bands:}
Our frequency-based token selection isolates a specific band bounded between the 10,000 and 40,000 rank thresholds. Although this specific window effectively avoids ubiquitous functional words and extreme low-frequency noise, it means that the impact on words outside this range is not explored.

\paragraph{Binary Trait of the WCS:}
A limitation of the WCS is its binary formulation of reachability ($\mathcal{R} \in \{0,1\}$). The metric functions as an analytical tool for absolute exclusion, highlighting moments where a sophisticated token sequence becomes mathematically impossible to generate. It does not capture cases of soft probability suppression, where a word remains technically inside the valid sampling set $\mathcal{V}_\theta$ but has its probability mass significantly reduced.

\paragraph{Static Context and Prefix Length Constraints:}
The Forced-Path Audit operates within a localized window bounded by a fixed prefix length of $L=256$ tokens. This provides sufficient context for localized syntactic evaluation but does not assess how long-range contextual dependencies interact with token probabilities.

\paragraph{Tokenizer Sub-Word Segmentation:}
The models evaluated utilize highly diverse sub-word tokenization vocabularies and therefore the same target word $w$ can be split into a differing number of sub-word fragments across different model families. The number of tokens determines the number of filtering steps that must be passed for the word to be reachable. This means that results across models are not fully comparable for some words.

\section{Acknowledgements}

This work was supported by the Agencia Estatal de Investigación (AEI) (doi:10.13039/501100011033) under Grant FUN4DATE (PID2022-136684OB-C22), by TUCAN6-CM (TEC-2024/COM460) funded by CM (ORDEN 5696/2024) and by European Union’s Horizon Europe research and innovation programme under project BRIDGE-AI (Grant 101299050). Access to the Gemini model was provided by the Google Cloud Research Credits program for the Gemini Academic Program.

\bibliographystyle{unsrt}
\bibliography{references}  

\clearpage

\appendix
\section{Appendix: Selected Word List and Aggregate Reachability}
\label{app:selected-words}

Table \ref{tab:selected-word-list} lists the 100 target words sampled from the $10\text{k}\text{--}40\text{k}$ frequency-rank band. Each word was paired with 10 PG-19 contexts, yielding 1,000 forced-path audit samples. The count column reports the source frequency count in millions. Within this appendix, more common selected words appear toward the beginning of the table, while rarer selected words appear toward the end. Thus, words such as \textit{offenders}, \textit{scattered}, and \textit{profitable} occupy the higher-frequency end of the selected band, whereas \textit{nodding}, \textit{unbounded}, \textit{saddened}, and \textit{precipitated} occupy the lower-frequency tail.

The selected words vary substantially in reachability even within the same frequency band. To summarize this variation, we aggregate reachability for each word across the evaluated models, contexts, temperatures, and sampler settings.

\begin{table}[h]
\scriptsize
\centering
\scriptsize
\caption{Selected target words from the $10\text{k}\text{--}40\text{k}$ frequency band. Frequency counts are shown in millions.}
\label{tab:selected-word-list}
\resizebox{\textwidth}{!}{%
\begin{tabular}{lrrlrrlrr}
\toprule
Word & Rank & Count & Word & Rank & Count & Word & Rank & Count \\
\midrule
offenders & 10104 & 4.98 & scattered & 10224 & 4.88 & profitable & 10437 & 4.73 \\
demon & 10618 & 4.60 & executing & 12138 & 3.70 & meanings & 12210 & 3.66 \\
crimson & 12536 & 3.52 & strangers & 13373 & 3.18 & smoked & 13540 & 3.12 \\
shocks & 13662 & 3.07 & badges & 13706 & 3.05 & averaged & 13767 & 3.02 \\
purity & 13814 & 3.01 & brewing & 13896 & 2.97 & supposedly & 14095 & 2.90 \\
excludes & 14161 & 2.88 & deliberately & 14211 & 2.86 & moderately & 14934 & 2.63 \\
disadvantage & 15341 & 2.50 & petitions & 15363 & 2.50 & horns & 15609 & 2.43 \\
cords & 15803 & 2.38 & ovarian & 15885 & 2.36 & acknowledges & 16036 & 2.32 \\
exceptionally & 16242 & 2.27 & recurrent & 16516 & 2.21 & parcels & 16573 & 2.20 \\
appealed & 16941 & 2.12 & surveyors & 16943 & 2.12 & utter & 17054 & 2.10 \\
lax & 17068 & 2.09 & inmate & 17298 & 2.05 & discomfort & 17321 & 2.04 \\
practicable & 17595 & 1.99 & buggy & 17771 & 1.96 & stare & 17937 & 1.93 \\
suction & 18223 & 1.88 & multiplied & 18754 & 1.79 & occult & 18859 & 1.77 \\
retiring & 19227 & 1.72 & tyranny & 19906 & 1.62 & jug & 20051 & 1.60 \\
friendships & 21108 & 1.46 & tak & 21484 & 1.41 & folly & 21891 & 1.37 \\
prosecuted & 22106 & 1.35 & denomination & 22154 & 1.34 & enumerated & 22308 & 1.32 \\
morphine & 22726 & 1.28 & pinned & 22805 & 1.28 & dubious & 23071 & 1.25 \\
arrears & 23918 & 1.18 & exhaustion & 24740 & 1.11 & bedside & 24974 & 1.09 \\
bleak & 25029 & 1.09 & undecided & 25269 & 1.07 & startling & 25270 & 1.07 \\
halves & 25288 & 1.07 & piers & 25427 & 1.06 & projecting & 25577 & 1.05 \\
guarding & 26209 & 1.01 & circulate & 26691 & 0.98 & sylvan & 27065 & 0.95 \\
reiterated & 27243 & 0.94 & moaning & 27785 & 0.91 & pronounce & 28483 & 0.87 \\
caprice & 28912 & 0.85 & dispositions & 29839 & 0.80 & ascend & 29929 & 0.80 \\
doubtless & 30020 & 0.80 & clutches & 30095 & 0.79 & dishonesty & 30443 & 0.78 \\
guise & 30656 & 0.77 & triumphant & 31354 & 0.74 & dormitory & 31439 & 0.74 \\
dictation & 32554 & 0.70 & fillet & 32578 & 0.69 & robbers & 32695 & 0.69 \\
roughness & 33513 & 0.66 & legions & 33784 & 0.65 & vulture & 34063 & 0.64 \\
feces & 34197 & 0.64 & manifests & 34319 & 0.64 & whirl & 34730 & 0.62 \\
scourge & 35321 & 0.61 & intolerable & 35437 & 0.60 & romances & 35582 & 0.60 \\
intimates & 35642 & 0.60 & apologized & 37027 & 0.56 & proclaiming & 37125 & 0.56 \\
volley & 37393 & 0.55 & bridle & 37428 & 0.55 & mattered & 37668 & 0.54 \\
murky & 38136 & 0.53 & embellished & 38237 & 0.53 & workmen & 38637 & 0.52 \\
nodding & 39571 & 0.50 & unbounded & 39731 & 0.49 & saddened & 39755 & 0.49 \\
precipitated & 39823 & 0.49 & & & & & & \\
\bottomrule
\end{tabular}%
}
\end{table}

\paragraph{Per-word reachability.}
For each target word, \textit{mean reachability} is the average fraction of evaluated model/sampler/context
conditions in which the word remained reachable. This provides a word-level view of which selected targets were
consistently preserved by the decoding process and which were more often pruned.

\begin{figure}[H]
    \centering
    \includegraphics[width=0.8\textwidth]{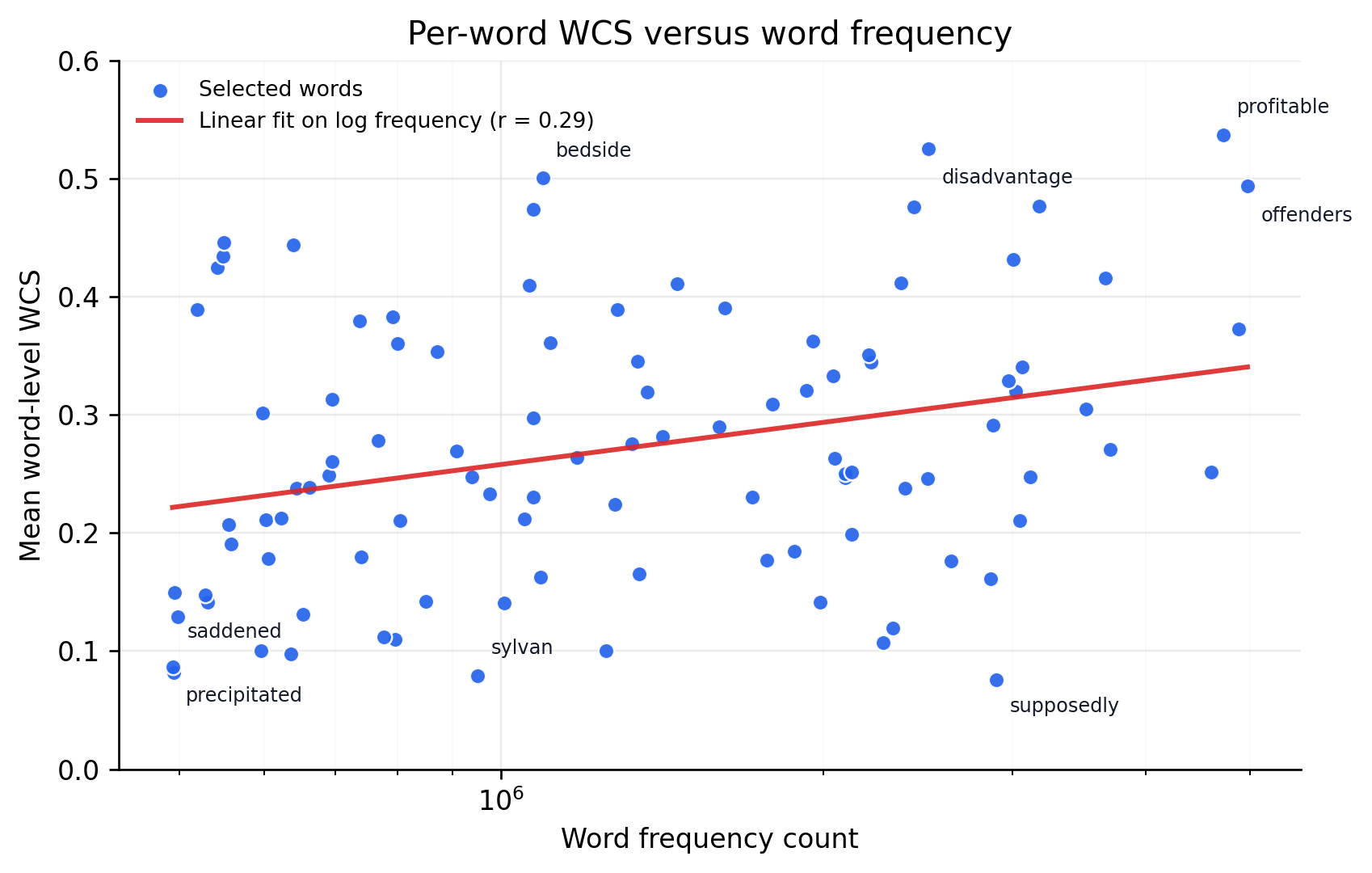}
    \caption{Mean word-level WCS versus source word frequency for the selected target words. The x-axis uses a
    logarithmic scale.}
    \label{fig:appendix-word-frequency-wcs}
\end{figure}

Figure \ref{fig:appendix-word-frequency-wcs} plots this mean word-level WCS against the source frequency count
for each selected word. The scatter shows that frequency alone does not explain reachability: some comparatively
frequent words remain difficult to reach, while some lower-frequency words are preserved more often. Using
log-transformed source frequency, the Pearson correlation with mean word-level WCS is weakly positive
($r=0.29$), indicating that corpus frequency alone explains only a limited portion of the observed variation in
lexical reachability.

The hardest words to reach were not simply the rarest words in the selected band. Several low-reachability words
came from the middle or upper part of the sampled frequency range, including \textit{supposedly} (rank 14,095; mean reachability 0.076), \textit{exceptionally} (rank 16,242; 0.107), and \textit{acknowledges} (rank 16,036; 0.119). The lowest-reachability group also included rarer words such as \textit{sylvan} (rank 27,065; 0.079), \textit{saddened} (rank 39,755; 0.082), and \textit{precipitated} (rank 39,823; 0.087). This indicates that lexical reachability is not determined by corpus frequency alone; it also depends on tokenizer segmentation, context, and model-specific probability structure.

Conversely, the easiest words to reach included both relatively frequent and lower-frequency targets. The highest-scoring words were \textit{profitable} (rank 10,437; mean reachability 0.537), \textit{disadvantage} (rank 15,341; 0.526), \textit{bedside} (rank 24,974; 0.501), and \textit{offenders} (rank 10,104; 0.494). Some lower-frequency words, such as \textit{volley} (rank 37,393; 0.446), \textit{feces} (rank 34,197; 0.444), \textit{bridle} (rank 37,428; 0.434), and \textit{workmen} (rank 38,637; 0.389), were also comparatively reachable. This reinforces that the WCS captures a decoding-level accessibility property rather than merely reproducing word-frequency rank.

\begin{table}[h]
\scriptsize
\centering
\caption{Examples of selected target words with low and high per-word reachability. Mean reachability is averaged across evaluated models, contexts, temperatures, and sampler settings.}
\label{tab:per-word-reachability}
\begin{tabular}{lrrlrr}
\toprule
\multicolumn{3}{c}{Harder to reach} & \multicolumn{3}{c}{Easier to reach} \\
\cmidrule(lr){1-3}\cmidrule(lr){4-6}
Word & Rank & Mean reachability & Word & Rank & Mean reachability \\
\midrule
supposedly & 14095 & 0.076 & profitable & 10437 & 0.537 \\
sylvan & 27065 & 0.079 & disadvantage & 15341 & 0.526 \\
saddened & 39755 & 0.082 & bedside & 24974 & 0.501 \\
precipitated & 39823 & 0.087 & offenders & 10104 & 0.494 \\
manifests & 34319 & 0.098 & strangers & 13373 & 0.477 \\
dubious & 23071 & 0.100 & horns & 15609 & 0.476 \\
intimates & 35642 & 0.100 & halves & 25288 & 0.474 \\
exceptionally & 16242 & 0.107 & volley & 37393 & 0.446 \\
doubtless & 30020 & 0.110 & feces & 34197 & 0.444 \\
dishonesty & 30443 & 0.112 & bridle & 37428 & 0.434 \\
\bottomrule
\end{tabular}
\end{table}

\end{document}